# Listen to Your Face: Inferring Facial Action Units from Audio Channel

Zibo Meng, *Student Member, IEEE,* Shizhong Han, *Student Member, IEEE,* and Yan Tong, *Member, IEEE*

**Abstract**—Extensive efforts have been devoted to recognizing facial action units (AUs). However, it is still challenging to recognize AUs from spontaneous facial displays especially when they are accompanied with speech. Different from all prior work that utilized visual observations for facial AU recognition, this paper presents a novel approach that recognizes speech-related AUs exclusively from audio signals based on the fact that facial activities are highly correlated with voice during speech. Specifically, dynamic and physiological relationships between AUs and phonemes are modeled through a continuous time Bayesian network (CTBN); then AU recognition is performed by probabilistic inference via the CTBN model.
A pilot audiovisual AU-coded database has been constructed to evaluate the proposed audio-based AU recognition framework. The database consists of a "clean" subset with frontal and neutral faces and a challenging subset collected with large head movements and occlusions. Experimental results on this database show that the proposed CTBN model achieves promising recognition performance for 7 speech-related AUs and outperforms the state-of-the-art visual-based methods especially for those AUs that are activated at low intensities or "hardly visible" in the visual channel. Furthermore, the CTBN model yields more impressive recognition performance on the challenging subset, where the visual-based approaches suffer significantly.

## 1 Introduction

FACIAL activity is one of the most powerful and natural means for human communication [1]. Extensive efforts have been devoted to facial activity analysis, most of which focused on recognizing six basic expressions, i.e., anger, disgust, happiness, fear, sadness, and surprise. To describe more complex facial activities, Facial Action Coding System (FACS) [2] defines a set of facial action units (AUs), each of which is anatomically related to the contraction of a set of facial muscles. An automatic facial AU recognition system is desired in many applications, such as human behavior analysis, interactive games, online learning, etc.

Facial AU recognition from static images or videos has received an increasing interest during the past decades as elaborated in the survey papers [3]–[5]. In spite of progress on posed facial displays and controlled image acquisition, recognition performance degrades significantly for spontaneous facial displays with free head movements, occlusions, and various illumination conditions [6]. More importantly, it is extremely challenging when recognizing AUs involved in speech production, since these AUs are usually activated at low intensities with subtle facial appearance/geometrical changes and often introduce ambiguity in detecting other co-occurring AUs [2], i.e., non-additive effects of AUs in a combination. For example, pronouncing a phoneme /p/ has two consecutive phases, i.e., *Stop* and *Aspiration* phases. As shown in Fig. 1(b), the lips are apart and the oral cavity between the teeth is visible in the *Aspiration* phase, based on which AU25 (lips part) and AU26 (jaw drop) can be detected from the image. Whereas, during the *Stop* phase as shown in Fig. 1(a), the lips are pressed together due to the activation of AU24 (lip presser).

• *Zibo Meng, Shizhong Han, and Yan Tong are with the Department of Computer Science and Engineering, University of South Carolina, Columbia, SC, 29208 USA.*
*E-mail: mengz, han38, tongy@**email.sc.edu***

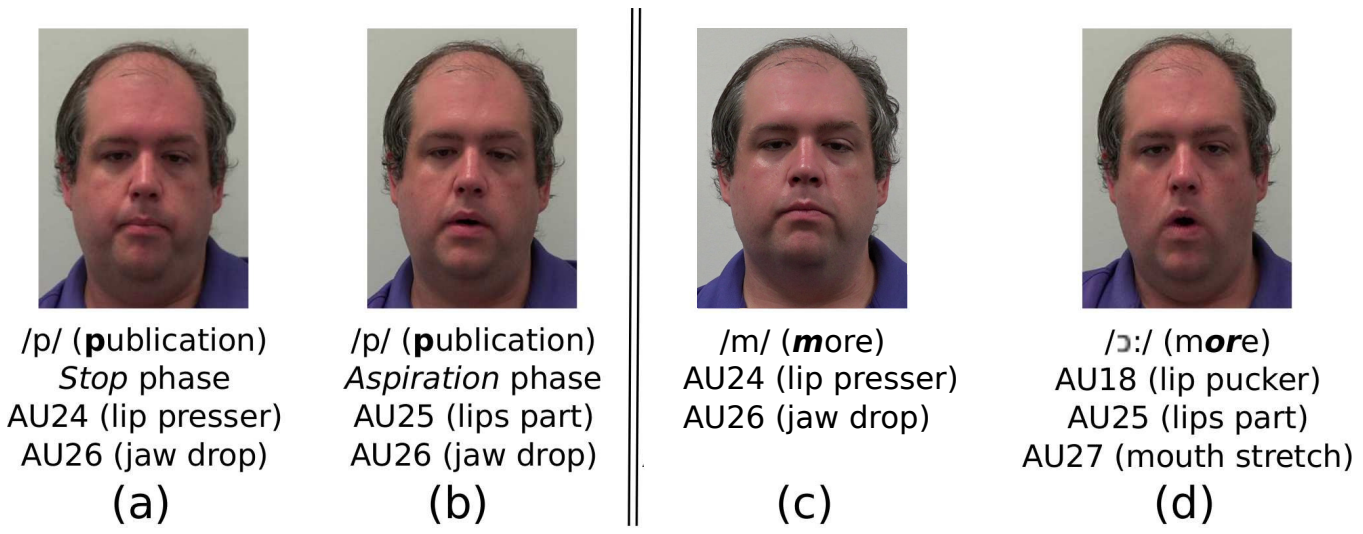

Fig. 1: Example images of speech-related facial behaviors, where different AUs are activated to pronounce sounds. Note non-additive effects of AUs co-occurring in a combinations in (a) and (d).

Since the oral cavity is occluded by the lips, AU26 is difficult to be detected from the visual channel. In another example, the oral cavity is partially occluded by the lips when producing /ɔ:/ in Fig. 1(d) due to the activation of AU18 (lip pucker). Hence, even the mandible is pulled down significantly, it is difficult to detect AU27 (mouth stretch) from Fig. 1(d).

These facial activities actually can be "heard", i.e., inferred from the information extracted from the audio channel. Facial AUs and voice are highly correlated in two ways. First, voice/speech has strong physiological relationships with some lower-face AUs such as AU24, AU26, and AU27, because jaw and lower-face muscular movements are the major mechanisms to produce differing sounds. In addition, eyebrow movements and fundamental frequency of voice have been found to be correlated during speech [7]. As demonstrated by the McGurk effect [8], there is a strong correlation between visual and audio information for speech perception. Second, both facial AUs and voice/speech convey human emotions in human communications. **Since the second type of relationships is emotion and context dependent, we will focus on studying the physiological relationships between** **lower-face AUs and speech, which are more objective and will generalize better to various contexts.**

In audiovisual automatic speech recognition (ASR), a *viseme* has been defined to represent facial muscle movements that can visually distinguish the sound [9]–[11]. Since some phonemes have similar facial appearance when produced, the mapping from phoneme to viseme is usually derived by statistical clustering [12]–[15], but without a universal agreement. Furthermore, the mapping is not always one-to-one because the number of visemes is usually less than the number of phonemes. For example, Neti et al. [14] clustered 44 phonemes into 13 visemes. However, one viseme may be produced by different AU or AU combinations or by a sequence of AU or AU combinations. For example, */p/* and */m/* are in the same cluster of bilabial consonants [14]. */p/* is produced by AU24 (lip presser) + AU26 (jaw drop) (Fig. 1a) followed by AU25 (lips part) + AU26 (jaw drop) (Fig. 1b); while */m/* is produced by AU24 (lip presser) + AU26 (jaw drop) as shown in Fig. 1c. Based on these observations, we proposed to directly study the relationships between facial AUs and phonemes rather than utilizing visemes as intermediate descriptors.

Specifically, a phoneme, which is the smallest phonetic unit of speech, is pronounced by activating a combination of AUs as illustrated in Fig. 1. Due to the variation in individual subjects, such relationships are stochastic. Furthermore, different combinations of AUs are responsible for sounding a phoneme at different phases as depicted in Fig. 1(a) and (b). Therefore, the dynamic dependencies between AUs and phonemes also undergo a temporal evolution rather than stationary.

Inspired by these, we proposed a novel approach to recognize speech-related AUs from speech by modeling and exploiting the dynamic and physiological relationships between AUs and phonemes through a Continuous Time Bayesian Network (CTBN) [16]. CTBNs are probabilistic graphical models proposed by Nodelman [16] to explicitly model the temporal evolutions over continuous time. CTBNs have been found in different applications, including users' presence and activities modeling [17], robot monitoring [18], sensor networks modeling [19], object tracking [20], host level network intrusion detection [21], dynamic system reliability modeling [22], social network dynamics learning [23], cardiogenic heart failure diagnosis and prediction [24], and gene network reconstruction [25].

Dynamic Bayesian networks are widely used dynamic models for modeling the dynamic relationships among random variables, and have been employed for modeling relationships among facial AUs in the visual channel [26], [27]. However, the dynamic events need to be discretized into discrete time points and thus, the relationships between them are modeled discontinuously. In addition, an alignment strategy should be employed to handle the difference in time scales and the time shift between the two signals. In contrast, considering AUs and phonemes as continuous dynamic events, the CTBN model can explicitly characterize the relationships between AUs and phonemes, and more importantly, model the temporal evolution of the relationships as a stochastic process over continuous time. Fig. 2 illustrates the proposed audio-based AU recognition system. During the training process (Fig. 2(a)), ground truth labels of AUs and phonemes are employed to learn the relationships between AUs and phonemes in a CTBN model. Furthermore, this model should also account for the uncertainty in speech recognition. For online AU recognition, as shown in Fig. 2(b), measurements of phonemes are obtained by automatic speech recognition and employed as evidence by the CTBN model; then AU recognition is performed by probabilistic inference over the CTBN model.

This work has three major contributions.

- The dynamic and physiological relationships between AUs and phonemes are theoretically and probabilistically modeled using a CTBN model.
- Instead of using low-level acoustic features, accurate phoneme measurements are employed benefiting from advanced speech recognition techniques.
- A pilot AU-coded audiovisual database is constructed to evaluate the proposed audio-based AU recognition framework and can be employed as a benchmark database for AU recognition.

The audiovisual AU-coded database consists of a "clean" subset with frontal and neutral faces and a challenging subset collected under unconstrained conditions with large head movements, occlusions from facial hair and accessories, and illumination changes. Experimental results on this database show that the proposed audio-based AU recognition framework achieves significant improvement in recognizing 7 speech-related AUs as compared to the state-of-the-art visual-based methods. The improvement is more impressive for those AUs that are activated at low intensities or "hardly visible" in the visual channel. More importantly, dramatic improvement has been achieved on the challenging subset: the average F1 score of the 7 speech-related AUs is almost doubled compared to those of the visual-based approaches.

## 2 Related Work

As elaborated in the survey papers [3]–[5], the current practice for facial AU recognition directly employs either spatial or temporal features, which are extracted from the visual channel, i.e., static images or videos, to capture the visual appearance or geometry changes caused by a specific AU.

### 2.1 Visual-based Approaches

The visual-based features utilized can be human-crafted and thus, general purpose. The features widely adopted in facial expression or facial AU recognition include magnitudes of a set of multiscale and multiorientation Gabor wavelets extracted either from the whole face region or at a few fiducial points [26]–[32], Haar wavelet features [32] considering the intensity difference of adjacent regions, and Scale Invariant Feature Transform (SIFT) features [33] extracted from a set of keypoints that are invariant to uniform scaling and orientation. Histograms of features extracted from a predefined facial grid have also been employed using Local Binary Patterns (LBPs) [6], [34], [35], Histograms of Oriented Gradients (HOG) [36], Local Phase Quantization (LPQ) features [37], and Local Gabor Binary Patterns (LGBP) [38], [39]. In addition, spatiotemporal extensions of the aforementioned 2D features such as LBP-TOP [40], LGBP-TOP [41], [42], LPQ-TOP [37], and dynamic Haar features [43], [44], which are usually calculated from three orthogonal planes, have been proposed to capture spatiotemporal changes.

In addition to the human-crafted feature representations, features can also be learned in a data-driven manner by sparse coding [45]–[51] or deep learning [52]–[64]. As an over-complete representation learned from given input, sparse coding can capture a wide range of variations that are not targeted to a specific application and has achieved promising results in facial expression recognition [47]–[51]. Non-negative Sparse Coding (NNSC) [65], integrating sparse coding and Non-negative Matrix Factorization (NMF) [66], has been adopted in facial expression recognition [67]–[70]. To learn representations that are more robust to the real-world applications [71], deep learning has been employed for facial expression recognition including deep belief network based approaches [54], [56], [57], [60] and convolutional neural network (CNN) based approaches [52], [53], [55], [58], [59], [61]–[64], [72]. Most of these deep-learning based methods took the whole face region as input and learned the high-level representations through a set of processing layers.

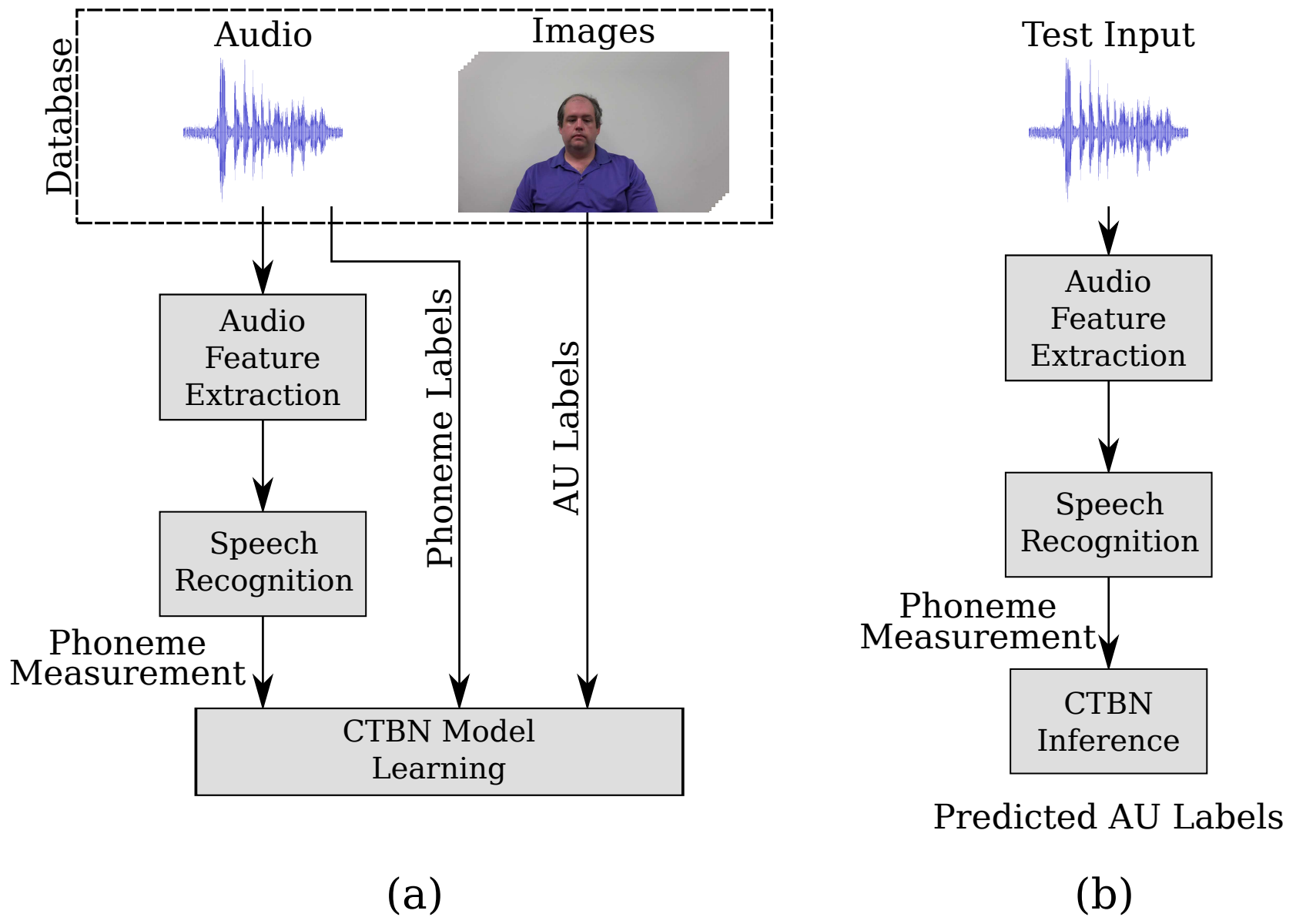

Fig. 2: The flowchart of the proposed audio-based AU recognition system: (a) an offline training process for CTBN model learning and (b) an online AU recognition process via probabilistic inference.

All the aforementioned visual-based approaches extract information, i.e., features, from the visual channel, and thus are challenged by imperfect image/video acquisition due to pose variations and occlusions and more importantly, by the non-additive effects as illustrated in Fig. 1 in recognizing speech-related AUs.

### 2.2 Audio-based Approaches

Recently, recognizing facial activities in the audio channel has been briefly studied in [74]–[76]. Lejan et al. [74] detected three types of facial activities, i.e. eyebrow movement, smiling, and head shaking, from audio signals. Based on the assumption that these facial activities are uncorrelated, different groups of low-level acoustic features are extracted independently for different facial activities, respectively. Then, artificial neural networks (ANNs) and RepTree are employed for classification. Ringeval et al. [75] utilized low-level acoustic feature sets, i.e. ComParE and GeMAPS, along with SVM and Long Short-Term Memory Recurrent Neural Network (LSTM-RNN) as classifiers to recognize facial AUs for emotion recognition. Most recently, our early work [76] employed a combination of audio features, i.e., Mel Frequency Cepstral Coefficents (MFCCs), and visual features (LBP features and CNN features) to improve recognition performance of speech-related AUs. However, all these aforementioned audio-based methods only utilize low-level acoustic features, which are susceptible to noise in the audio channel, whereas semantic and dynamic relationships between audio and visual channels are ignored.

In contrast, the proposed audio-based AU recognition framework employs information extracted solely from the audio channel to recognize speech-related AUs. Instead of directly utilizing low-level acoustic features, high-level audio information, i.e., phonemes, is employed by taking advantage of the advanced speech recognition techniques. Moreover, a CTBN-based probabilistic framework is developed to systematically model the dynamic and physiological relationships between AUs and phonemes over continuous time.

## 3 Audio-based Facial Action Modeling

### 3.1 Phoneme-AU Relationship Analysis

A phoneme is defined as the smallest phonetic unit in a language. In this work, a set of phonemes defined by Carnegie Mellon University Pronouncing Dictionary (CMUdict) [77] is employed, which is a machine-friendly pronunciation dictionary designed for speech recognition, where 39 phonemes are used for describing North American English words. The 39 phonemes defined by CMUdict, along with sample words in parenthesis, are as follows: *AA* (**o**dd), *AE* (**a**t), *AH* (h**u**t), *AO* (**aw**ful), *AW* (c**ow**), *AY* (h**i**de), *B* (**b**e), *CH* (**ch**eese), *D* (**d**ee), *DH* (**th**ee), *EH* (**E**d), *ER* (h**ur**t), *EY* (**a**te), *F* (**f**ee), *G* (**g**reen), *HH* (**h**e), *IH* (**i**t), *IY* (**ea**t), *JH* (**g**ee), *K* (**k**ey), *L* (**l**ee), *M* (**m**e), *N* (**kn**ee), *NG* (pi**ng**), *OW* (**oa**t), *OY* (t**oy**), *P* (**p**ee), *R* (**r**ead), *S* (**s**ea), *SH* (**sh**e), *T* (**t**ea), *TH* (**th**eta), *UH* (h**oo**d), *UW* (t**wo**), *V* (**v**ee), *W* (**w**e), *Y* (**y**ield), *Z* (**z**ee), *ZH* (sei**z**ure) [77].

Since each phoneme is anatomically related to a specific set of jaw and lower facial muscular movements, there are strong physiological relationships between the speech-related AUs and phonemes. Taking the word *gooey* for instance, a combination of AU25 (lip part) and AU26 (jaw drop) is first activated to produce *G* (***g**ooey*) (Fig. 3a). Then, AU18 (lip pucker), AU25, and AU26 are activated together to sound *UW* (*g**oo**ey*) (Fig. 3b). Finally, AU25 and AU26 are responsible for producing *IY* (*goo**ey***) (Fig. 3c).

Furthermore, these relationships also undergo a temporal evolution rather than stationary. *There are two types of temporal dependencies between AUs and phonemes*. First, a phoneme is produced by a combination of AUs as shown in Fig. 3. The probabilities of the AUs being activated increase prior to voicing

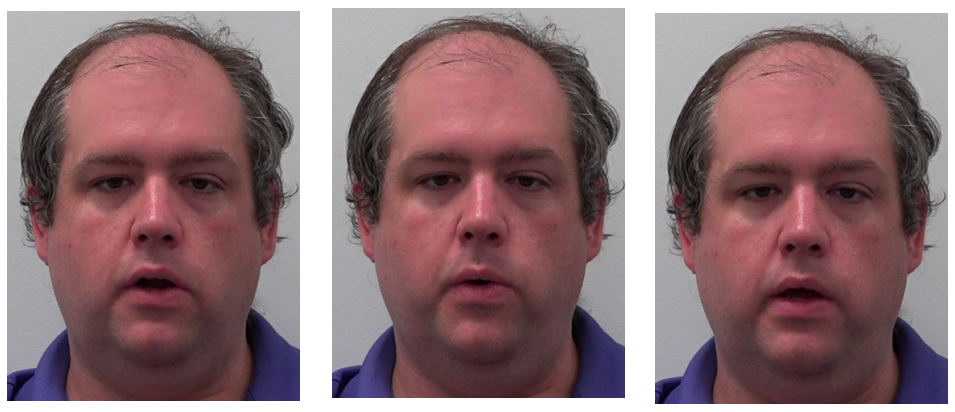

(a) G(***g***ooey) AU25+26 (b) UW(g***oo***ey) AU18+25+26 (c) IY(goo***ey***) AU25+26

Fig. 3: Examples of physiological relationships between phonemes and AUs. To pronounce a word *gooey* , different combinations of AUs are activated sequentially. (a) AU25 (lip part) and AU26 (jaw drop) are responsible for producing *G* (***g****ooey*); (b) AU18 (lip pucker), AU25, and AU26 are activated to pronounce *UW* (*g****oo****ey*); and (c) AU25 and AU26 are activated to sound *IY* (*goo****ey***).

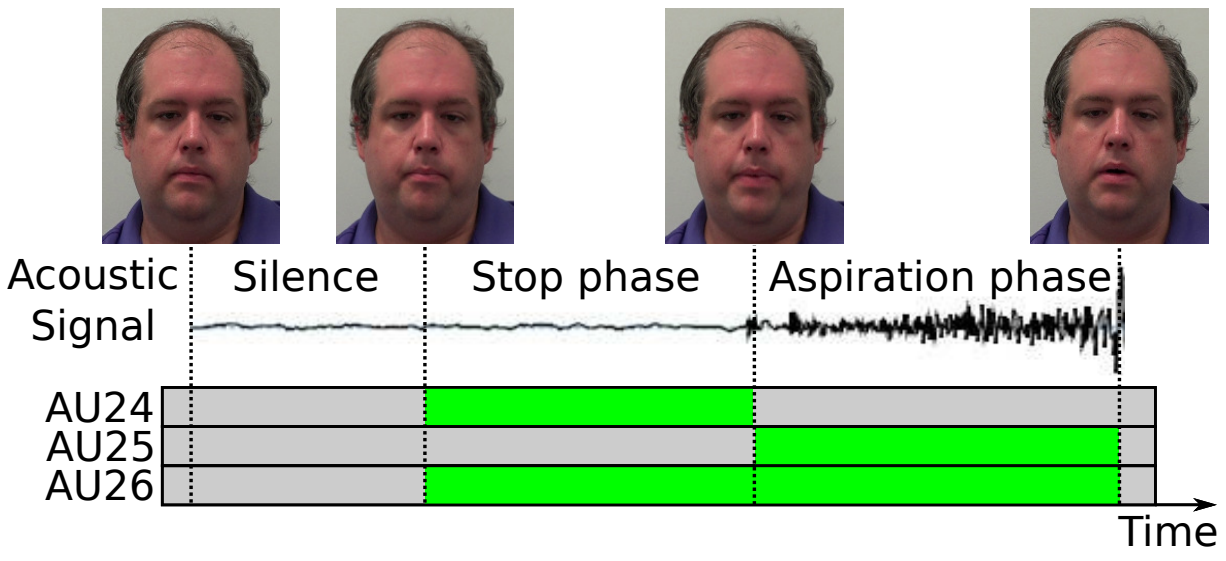

Fig. 4: Illustration of the dynamic relationships between AUs and phonemes while producing *P*. Specifically, AU24 (lip presser) and AU26 (jaw drop) are activated in the first phase, i.e., the Stop phase, while AU25 (lips part) and AU26 are activated in the second phase, i.e., the Aspiration phase. The activated AUs are denoted by green bars with a diagonal line pattern; while the inactivated AUs are denoted by grey bars. Best viewed in color.

the phoneme and reach an apex when the sound is fully emitted, and then decrease while preparing to voice the next phoneme.

Second, different combinations of AUs are responsible for producing a single phoneme at different phases. For example, as illustrated in Fig. 4, the phoneme *P* in the word *cha**p**s* has two consecutive phases, i.e., *Stop* and *Aspiration* phases. During the *Stop* phase, AU24 (lip presser) is activated as lips are pressed together to hold the breath without making sound [78], when the upper and lower teeth are apart indicating the presence of AU26 (jaw drop). In the *Aspiration* phase, the lips are apart by activating AU25 and releasing AU24 to release the breath with an audible explosive sound [78]. Thus, AU24 and AU26 are activated before the sound is heard, and AU24 is released as soon as the sound is made when AU25 is activated.

Note that these dynamic and physiological relationships are stochastic and vary among individual subjects and different words. For example, according to Phonetics [78], AU20 (lip stretcher) is required to produce *AE* in *ch**a**ps*. However, some subjects may not activate AU20 as observed in our audiovisual dataset. In addition, the duration of the *Stop* phase of *P* or *B* varies across different subjects and different words. In addition, the AUs are usually activated slightly before the phoneme is produced [79]. Therefore, we employ a probabilistic framework, a CTBN [16] in particular, to explicitly model the dynamic relationships between phonemes and AUs over continuous time.

### 3.2 Modeling Phoneme-AU Relationships by a CTBN

A CTBN is a directed, possibly cyclic, graphical model [16], which consists of an initial distribution specified as a Bayesian network and a set of random variables. As shown in Fig. 5, a CTBN model is employed to capture the dynamic and physiological relationships between AUs and phonemes as well as the measurement uncertainty in speech recognition. There are two types of nodes in the model: the *unshaded nodes* represent hidden nodes, whose states should be inferred through the model; whereas the *shaded node* denotes the measurement node, whose states can be observed and used as evidence for inference.

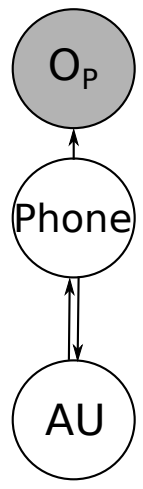

Fig. 5: A CTBN model for audio-based AU recognition.

Specifically, the phoneme node denoted by "**Phone**" has 29 states, i.e., 28 phonemes in the audiovisual dataset and one silence state, and is employed to model the dynamics of phonemes: durations of phonemes and transitions between phonemes. A measurement node denoted as "$\mathbf{O}_p$" with 29 states is used to represent the phoneme measurement obtained by speech recognition. The directed link between "**Phone**" and "$\mathbf{O}_p$" represents the measurement uncertainty in speech recognition, e.g., misdetection and temporal misalignment.

Based on the study in [26], there are semantic and dynamic relationships among AUs. In this work, AUs often occur in combinations to produce sounds. However, the CTBN model has an assumption that no two variables change at exactly the same time, which, unfortunately, is not held in this application, where two or even more different AUs can change simultaneously. For example, AU25 is activated at the same time when AU24 is released as illustrated in Fig 3.

Instead of using 7 separate nodes for 7 speech-related AUs, respectively, a single "**AU**" node is employed to model the joint distributions of all speech related AUs. Since each AU can be at one of "absence" or "presence" status, "**AU**" has $2^7 = 128$ states, each of which is corresponding to one combination of 7 AUs. For example, the state $0$ of the "**AU**" node represented by a binary number "0000000", means no AU is activated, while the state $1$ with a binary number "0000001", means only AU27 is present. This way, the relationships between all AUs are naturally modeled without learning the CTBN structure. The directed links between "**AU**" and "**Phone**" capture the dynamic and physiological relationships between them.

### 3.3 Model Parameterization

In a CTBN, each node, e.g., "**Phone**" and "**AU**" in this work, evolves as a Markov process, whose dynamics is described by a set of transition intensity matrices, called conditional intensity matrices (CIMs) denoted by $\mathbf{Q}$, in which the transition intensity values are determined by the instantiations of parent node(s).

#### 3.3.1 Model Parameterization for "Phone"

The directed link from "**AU**" to "**Phone**" represents the relationships that AUs are activated prior to pronounce a phoneme and thus, the dynamic of "**Phone**" is based on the instantiations of "**AU**". Given the $k^{th}$ state of "**AU**" denoted as $a_k$, $k = 0, \cdots, 127$, the CIM for "**Phone**", a $29 \times 29$ matrix denoted as $\mathbf{Q}_{\textbf{Phone}|\textbf{AU}=a_k}$, is defined as follows:

$$\mathbf{Q}_{\textbf{Phone}|\textbf{AU}=a_k} = \begin{bmatrix} -q_0^{ph|a_k} & q_{0,1}^{ph|a_k} & \cdots & q_{0,28}^{ph|a_k} \\ q_{1,0}^{ph|a_k} & -q_1^{ph|a_k} & \cdots & q_{1,28}^{ph|a_k} \\ \vdots & \vdots & \ddots & \vdots \\ q_{28,0}^{ph|a_k} & q_{28,1}^{ph|a_k} & \cdots & -q_{28}^{ph|a_k} \end{bmatrix} \quad (1)$$

where $q_i^{ph|a_k}$ denotes the conditional intensity value when "**Phone**" remains at its $i^{th}$ state denoted by $ph_i$, $i = 0, \cdots, 28$, given $\mathbf{AU} = a_k$; $q_{i,j}^{ph|a_k}$ ($j = 0, \cdots, 28$ and $j \neq i$) denotes the conditional intensity value when "**Phone**" transitions from its $i^{th}$ state to its $j^{th}$ state, given $\mathbf{AU} = a_k$; and $q_i^{ph|a_k} = \sum_{j \neq i} q_{i,j}^{ph|a_k}$.

Based on Eq. 1, the dynamics of "**Phone**" may change following the state of "**AU**". For example, if "**AU**" is at its $a_0$ state, the dynamics of "**Phone**" will be controlled by its CIM $\mathbf{Q}_{\textbf{Phone}|\textbf{AU}=a_0}$; while the intensity matrix $\mathbf{Q}_{\textbf{Phone}|\textbf{AU}=a_1}$ will be employed after "**AU**" transitions to its $a_1$ state.

Given the initial states of "**Phone**" and "**AU**" at time $t = 0$ (**Phone** $= ph_i$ and **AU** $= a_k$), the probability of **Phone** remaining at its initial state $ph_i$ is specified by the probability density function as [16]:

$$f(t) = q_i^{ph|a_k} e^{-q_i^{ph|a_k} t}, \quad t \geq 0 \quad (2)$$

Then, the expected time of transition of "**Phone**", i.e., leaving from the $i^{th}$ state to any of the other states, can be computed as $\frac{1}{q_i^{ph|a_k}}$. When transition occurs, "**Phone**" transitions from its $i^{th}$ state to its $j^{th}$ state with probability denoted by $\theta_{i,j|a_k} = \frac{q_{i,j}^{ph|a_k}}{q_i^{ph|a_k}}$ [16].

### 3.3.2 Model Parameterization for "AU" and "$\mathbf{O}_p$"

The state of "**AU**" may also change according to the state of "**Phone**". Following the previous example of producing a phoneme *P*, the probability of AU24 (lip presser) should decrease rapidly if the sound is emitted in the *Aspiration* phase. Such relationships can be captured by a directed link from "**Phone**" to "**AU**". Then, the CIM of "**AU**" given the $i^{th}$ state of "**Phone**" is denoted by $\mathbf{Q}_{\textbf{AU}|\textbf{Phone}=ph_i}$ and can be defined similarly as Eq. 1. Likewise, the CIM of "$\mathbf{O}_p$" given the $i^{th}$ state of "**Phone**" ($\mathbf{Q}_{\mathbf{O}_p|\textbf{Phone}=ph_i}$) captures the measurement uncertainty of speech recognition and is defined similarly as Eq. 1.

## 3.4 Parameter Estimation

The model parameters of a CTBN include the initial distribution $Pr_0$ specified by a Bayesian network, the structure of CTBN, and the CIMs. The initial distribution $Pr_0$ can be estimated given the groundtruth AU and phoneme labels of the first frames of all sequences. It becomes less important in the context of CTBN inference and learning when we assume the model is irreducible, especially when the time range becomes significantly large [21]. Thus, as the CTBN model structure is given as shown in Fig. 5, the model parameters we should learn are the expected time of transitions, i.e., $\frac{1}{q_i^{ph|a_k}}$, and the transition probabilities, i.e., $\theta_{i,j|a_k}$. In this work, the groundtruth AU labels and the phoneme labels are manually annotated, and thus the training data $\mathcal{D}$ is complete, i.e., for each time point along each trajectory, the instantiation of all variables is known. Then, we can estimate the parameters of a CTBN efficiently using Maximum Likelihood estimation (MLE) [80]. In particular, the likelihood function can be factorized as the product of a set of local likelihood functions as below:

$$\begin{aligned} L(\mathbf{q}, \boldsymbol{\theta} : \mathcal{D}) &= \prod_{X \in \mathbf{X}} L_X(\mathbf{q}_{X|\mathbf{v}_X}, \boldsymbol{\theta}_{X|\mathbf{v}_X} : \mathcal{D}) \\ &= \prod_{X \in \mathbf{X}} L_X(\mathbf{q}_{X|\mathbf{v}_X} : \mathcal{D}) L_X(\boldsymbol{\theta}_{X|\mathbf{v}_X} : \mathcal{D}) \end{aligned} \quad (3)$$

where $\mathbf{X}$ consists of all random variables in the CTBN, i.e., "**AU**", "**Phone**", and "$\mathbf{O}_p$" in this work; $X \in \mathbf{X}$ is a random variable with $M$ states and has a set of parent nodes denoted by $\mathbf{V}_X$. $\mathbf{q}_{X|\mathbf{v}_X}$ is a set of parameters characterizing the expected time of transition from the current state of $X$ to any of the other states given its parent nodes $\mathbf{V}_X$, i.e., the diagonal elements of $\mathbf{Q}_{X|\mathbf{v}_X}$; and $\boldsymbol{\theta}_{X|\mathbf{v}_X}$ represents the transition probabilities of $X$ given its parent nodes $\mathbf{V}_X$, i.e., the off-diagonal elements of $\mathbf{Q}_{X|\mathbf{v}_X}$.

Given an instantiation of the parent nodes, i.e., $\mathbf{V}_X = \mathbf{v}_X$, the sufficient statistics are $T[x_i|\mathbf{v}_X]$ representing the total length of time that $X$ stays at the state $x_i$ and $N[x_i, x_j|\mathbf{v}_X]$ representing the number of transitions of $X$ from the state $x_i$ to the state $x_j$. With the sufficient statistics, $L_X(\mathbf{q}_{X|\mathbf{v}_X} : \mathcal{D})$ and $L_X(\boldsymbol{\theta}_{X|\mathbf{v}_X} : \mathcal{D})$ in Eq. 3 can be calculated as follows [80],

$$\begin{aligned} &L_X(\mathbf{q}_{X|\mathbf{v}_X} : \mathcal{D}) \\ &\quad = \prod_{\mathbf{v}_X} \prod_{i \in M} (q_i^{X|\mathbf{v}_X})^{N[x_i|\mathbf{v}_X]} \exp\left(-q_i^{X|\mathbf{v}_X} T[x_i|\mathbf{v}_X]\right) \end{aligned} \quad (4)$$

where $q_i^{X|\mathbf{v}_X}$ is the $i^{th}$ diagonal element in the CIM of $X$ given an instantiation of its parent nodes ($\mathbf{Q}_{X|\mathbf{v}_X}$, referring to Eq. 1); and $N[x_i|\mathbf{v}_X] = \sum_{j \in M, j \neq i} N[x_i, x_j|\mathbf{v}_X]$ represents the total number of transitions leaving from the state $x_i$.

$$L_X(\boldsymbol{\theta}_{X|\mathbf{v}_X} : \mathcal{D}) = \prod_{\mathbf{v}_X} \prod_{i \in M} \prod_{j \in M, j \neq i} (\theta_{i,j|\mathbf{v}_X})^{N[x_i, x_j|\mathbf{v}_X]} \quad (5)$$

where $\theta_{i,j|\mathbf{v}_X} = \frac{q_{i,j}^{X|\mathbf{v}_X}}{q_i^{X|\mathbf{v}_X}}$ represents the transition probability from the $i^{th}$ state of $X$ to the $j^{th}$ state, given an instantiation of its parent nodes $\mathbf{v}_X$.

By substituting Eq. 4 and Eq. 5 into Eq. 3, the log-likelihood for $X$ can be obtained as below

$$\begin{aligned} &\ell_X(\mathbf{q}_{X|\mathbf{v}_X}, \boldsymbol{\theta}_{X|\mathbf{v}_X} : \mathcal{D}) = \ell_X(\mathbf{q}_{X|\mathbf{v}_X} : \mathcal{D}) + \ell_X(\boldsymbol{\theta}_{X|\mathbf{v}_X} : \mathcal{D}) \\ &= \sum_{\mathbf{v}_X} \sum_{i \in M} N[x_i|\mathbf{v}_X] \ln(q_i^{X|\mathbf{v}_X}) - q_i^{X|\mathbf{v}_X} T[x_i|\mathbf{v}_X] \\ &+ \sum_{\mathbf{v}_X} \sum_{i \in M} \sum_{j \in M, j \neq i} N[x_i, x_j|\mathbf{v}_X] \ln \theta_{i,j|\mathbf{v}_X} \end{aligned} \quad (6)$$

By maximizing Eq. 6, the model parameters can be estimated as follows [81]:

$$\hat{q}_i^{X|\mathbf{v}_X} = \frac{N[x_i|\mathbf{v}_X]}{T[x_i|\mathbf{v}_X]} \quad (7)$$

$$\hat{\theta}_{i,j|\mathbf{v}_X} = \frac{N[x_i, x_j|\mathbf{v}_X]}{N[x_i|\mathbf{v}_X]} \quad (8)$$

### 3.5 Phoneme Measurements Acquisition

In this work, a state-of-the-art speech recognition approach, i.e., Kaldi toolkit [82], is employed to obtain the phoneme measurements. In particular, 13-dimensional MFCC features [83] are first extracted, based on which, Kaldi is used to produce word-level speech recognition results, which are further aligned into phoneme-level segments. These phoneme-level segments are then fed into the CTBN model as the evidence. Note that, the evidence is given as a continuous event and the gaps between two successive phonemes are considered as silence.

### 3.6 AU Recognition via CTBN Inference

Given the fully observed evidence, i.e., phoneme measurements denoted by $\mathbf{E}_p$, and a prior distribution, $Pr_0$, over the variables at time $t_0$, AU recognition is performed by estimating the posterior probability $Pr(\mathbf{AU}|\mathbf{E}_p)$ via the CTBN model. Exact inference can be performed by flattening all CIMs into a single intensity matrix $\mathbf{Q}$ using amalgamation, which will be treated as a homogeneous Markov process [16], where the intensity values in $\mathbf{Q}$ stay the same over time. However, exact inference is infeasible for this work as the state space grows exponentially large as the number of variables increases. In this work, we employ auxiliary Gibbs sampling [84], which takes a Markov Chain Monte Carlo (MCMC) approach to estimate the distribution given evidence, implemented in the CTBN reasoning and learning engine (CTBN-rle) [85] to perform CTBN inference.

Since the state of the "**AU**" node corresponds to the joint states of 7 speech-related AUs, the inference results would be the joint probability of those AUs. Then, the posterior probability of a target AU given the evidence can be obtained by marginalizing out all the other AUs. Optimal states of the target AUs can be estimated by maximizing the posterior probability.

## 4 Experimental Results

To demonstrate the effectiveness of the proposed approach, a pilot audiovisual database is constructed, on which the proposed method, a state-of-the-art visual-based method, and a set of baseline experiments are conducted.

### 4.1 Audiovisual Dataset

As far as we know, all the publicly available AU-coded datasets only provide visual information. Moreover, all speech-related AUs are either coded using a uniform label, i.e. AD50 [6], or completely ignored [39], during speech. Thus, in order to learn the dynamic and physiological relationships between AUs and phonemes, as well as to evaluate the proposed audio-based facial AU recognition framework, we constructed a pilot audiovisual database consisting of two subsets, i.e. a clean subset and a challenging subset. Fig. 6 illustrates example images of the speech-related AUs in the audiovisual database.

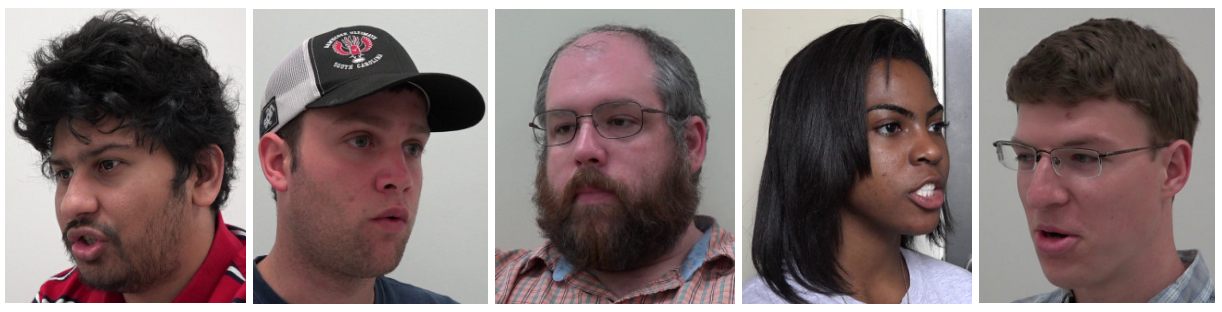

Fig. 7: Example images in the challenging subset collected from different illuminations, varying view angles, and with occlusions by glasses, caps, or facial hairs.

There are a total of 13 subjects in the audiovisual database, where 2 subjects appear in both the clean and challenging subsets. All the videos in this database were recorded at 59.94 frames per second at a spatial resolution of $1920 \times 1080$ with a bit-depth of 8 bits; and the audio signals were recorded at 48kHz with 16 bits. The statistics, i.e., the numbers of occurrences, of the speech-related AUs in the clean and challenging subsets are reported in Table. 1, respectively.

TABLE 1: Statistics of the speech-related AUs in the audiovisual database.

| Subsets | AU18 | AU20 | AU22 | AU24 | AU25 | AU26 | AU27 | Total Frames |
|---|---|---|---|---|---|---|---|---|
| Clean | 7,014 | 1,375 | 4,275 | 2,105 | 25,092 | 18,280 | 4,444 | 34,622 |
| Challenging | 4,118 | 1,230 | 3,396 | 1,373 | 17,554 | 11,830 | 3,242 | 23,274 |

In the clean subset, videos were collected from 9 subjects covering different races, ages, and genders. It consists of 12 words [1], which contain 28 phonemes and the most representative relationships between AUs and phonemes. Other phonemes are generally produced by AU25 (lips part) and AU26 (jaw drop), which represents the most common facial AU combination observed during speech. For example, even though there are two different phases to pronounce *T* in ***t***ea or *K* in ***k***ey, the same combination of facial AUs, i.e. AU25+AU26, is activated in both phases with tongue movements. Each subject was asked to speak the selected 12 words individually, each of which was repeated 5 times. In addition, all subjects were required to keep a neutral face during data collection to ensure all the facial activities are only caused by speech.

Videos in the challenging subset were collected from 6 subjects covering different races and genders speaking the same words for 5 times as those in the clean set. As illustrated in Fig. 7, the subjects were free to display any expressions on their face during speech and were not necessary to show neutral face before and after speaking the word, and there are occlusions on the face region caused by glasses, caps, and facial hairs, introducing challenges to AU recognition from the visual channel. Instead of being recorded from the frontal view, videos were collected mostly from the sideviews with free head movements. In addition, illumination sources were at different locations to the subjects, and some videos were collected under low illumination. Moreover, the videos were collected under unconstrained conditions with background noise and the microphone was mounted at different locations to the subjects, both of which introduced challenges to the audio channel.

1. The 12 words including "beige","chaps","cowboy","Eurasian", "gooey","hue","joined","more","patch","queen", "she", and "waters" were selected from English phonetic pangrams (http://www.liquisearch.com/list_of_pangrams/english_phonetic_pangrams) that consists of all the phonemes at least once in 53 words.

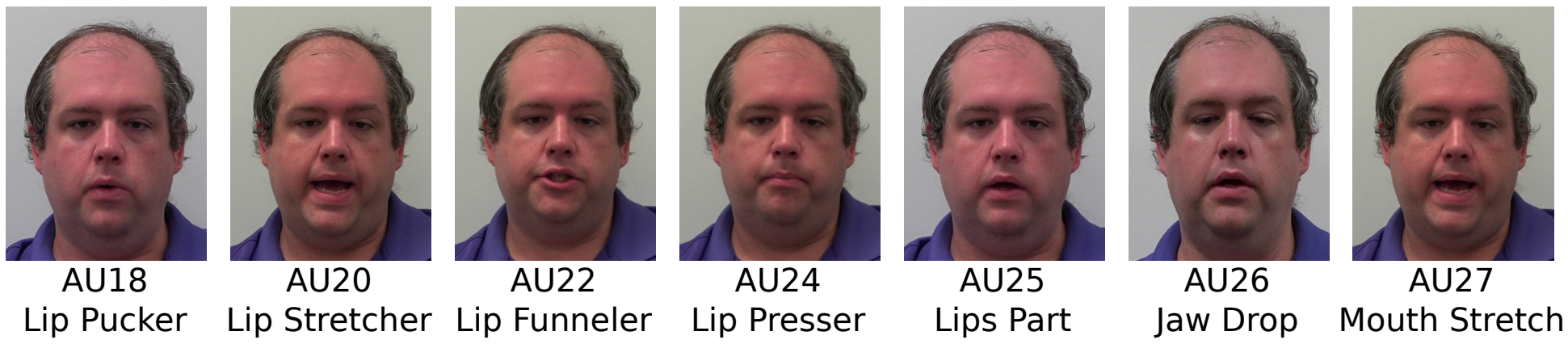

Fig. 6: Example images of the speech related AUs in the audiovisual database.

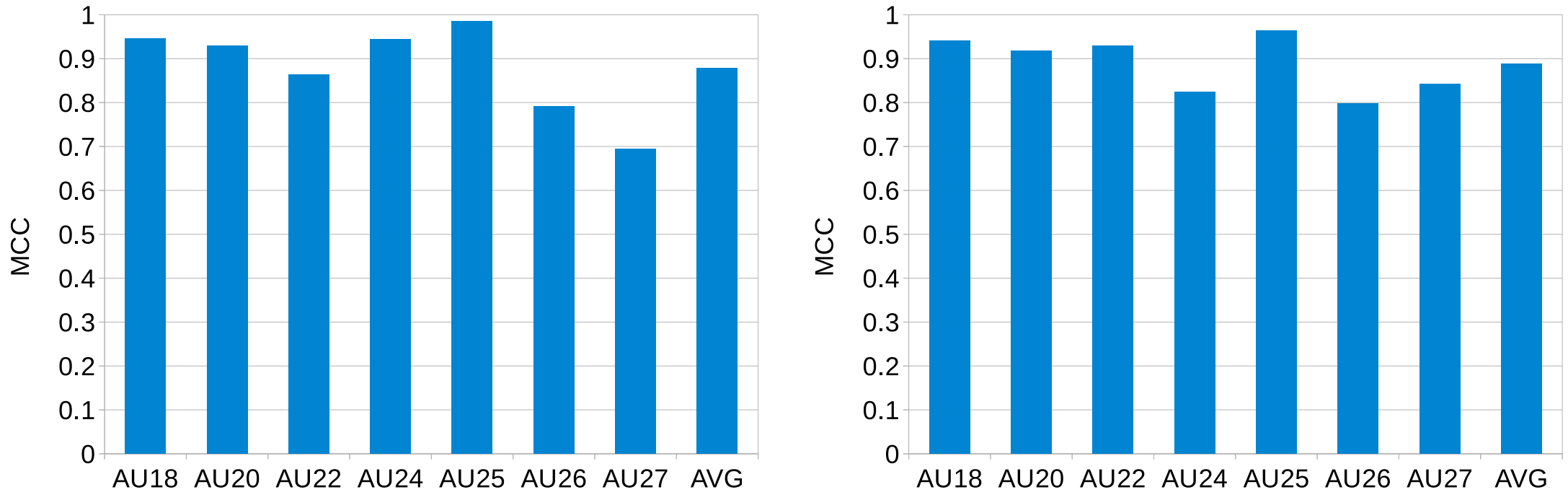

(a) the clean subset (b) the challenging subset

Fig. 8: Inter-coder reliability measured by MCC for the 7 speech-related facial AUs on (a) the clean subset and (b) the challenging subset.

Groundtruth phoneme segments and AU labels were recorded in the database. Specifically, the utterances were transcribed using the Penn Phonetics Lab Forced Aligner (p2fa) [86], which takes an audio file along with its corresponding transcript file as input and produces a Praat [87] TextGrid file containing the phoneme segments. 7 speech-related AUs, i.e. AU18, AU20, AU22, AU24, AU25, AU26, and AU27, as shown in Fig. 6, were frame-by-frame labeled manually by two certified FACS coders. Following the settings in [39], most of the data has been labeled only by one labeler, while roughly 10% of the data was labeled by two coders independently to estimate inter-coder reliability measured by Matthews Correlation Coefficient (MCC) [88]. When the visual observation was impeded by occlusions, the AU coders labeled AUs according to Phonetics [78] as well as self-reports from the subjects to achieve a good inter-coder agreement. As illustrated in Fig. 8, the MCC for each AU ranges from 0.69 for AU27 to 0.98 for AU25 and has an average of 0.88 on the clean subset (Fig. 8(a)), and ranges from 0.80 for AU26 to 0.96 for AU25 on the challenging subset (Fig. 8(b)), which indicates strong to very strong inter-coder reliability of AU annotation.

### 4.2 Baseline Methods for Comparison

To demonstrate the effectiveness of the proposed audio-based AU recognition framework, we compared the proposed method, denoted as *CTBN*, with seven baseline methods on the AU-coded audiovisual database.

**Ada-LBP:** The first visual-based baseline method, denoted as *Ada-LBP* [89], employs histogram of LBP features, which have been shown to be effective in facial AU recognition. Specifically, face regions across different facial images are aligned to remove the scale and positional variance based on a face and eye detector and then cropped to $96 \times 64$ for preprocessing purposes [2]. Then, the cropped face region is divided with a $7 \times 7$ grid, from each subregion of which, LBP histograms with 59 bins are extracted. AdaBoost is employed to select the most discriminative features, which are used to construct a strong classifier for each AU.

2. All the visual-based baseline methods employed the same preprocessing strategy. Except the IB-CNN-LIP, which employed a $96 \times 96$ face region, all methods used a $96 \times 64$ face region.

**Ada-LPQ:** The second visual-based baseline method, denoted as *Ada-LPQ* [37], employs histogram of LPQ features. Specifically, the face region is divided into $7 \times 7$ grid, from each of which, LPQ histograms with 256 bins are extracted. Similar to the *Ada-LBP*, AdaBoost is employed for feature selection and classifier construction for each AU.

**SVM-LGBP:** The third visual-based baseline method, denoted as *SVM-LGBP*, employed histogram of LGBP features [38], [39]. Particularly, the face region is convolved with 18 Gabor filters, i.e. three wavelengths $\lambda = \{3, 6.3, 13.23\}$ and six orientations $\theta = \{0, \frac{\pi}{6}, \frac{\pi}{3}, \frac{\pi}{2}, \frac{2\pi}{3}, \frac{5\pi}{6}\}$, with a phase offset $\psi = 0$, a standard deviation of the Gaussian envelope $\sigma = \frac{5\pi}{36}$, and a spatial aspect ratio $\gamma = 1$, which results in 18 Gabor magnitude response maps. Each of the response maps is divided into a $7 \times 7$ grid, from each of which, LBP histograms with 59 bins are extracted and concatenated as LGBP features. For each AU, AdaBoost is employed to select 400 LGBP features, which are employed to train an SVM classifier.

**IB-CNN-LIP:** The fourth baseline method, denoted as *IB-CNN-LIP*, employed a deep learning based model, i.e. Incremental Boosting Convolutional Neural Network (IB-CNN) [72] for facial AU recognition. Since only the lower-part of the face is responsible for producing the speech-related AUs, the aligned and cropped lip region along with the landmarks on lips are employed in a two-stream IB-CNN to learn both appearance and geometry information for each target AU.

**Ada-MFCC:** The fifth baseline method, denoted as *Ada-MFCC*, employs low-level audio features, i.e., 13-dimensional MFCC features, extracted from the audio channel. Since there is a random shift between the MFCC and the video frames, a cubic spline interpolation method is employed to synchronize MFCC features with the image frames [76]. In addition, 7 frames of the MFCC features, i.e. 3 frames before and after the current frame along with the current one, are concatenated as the final MFCC

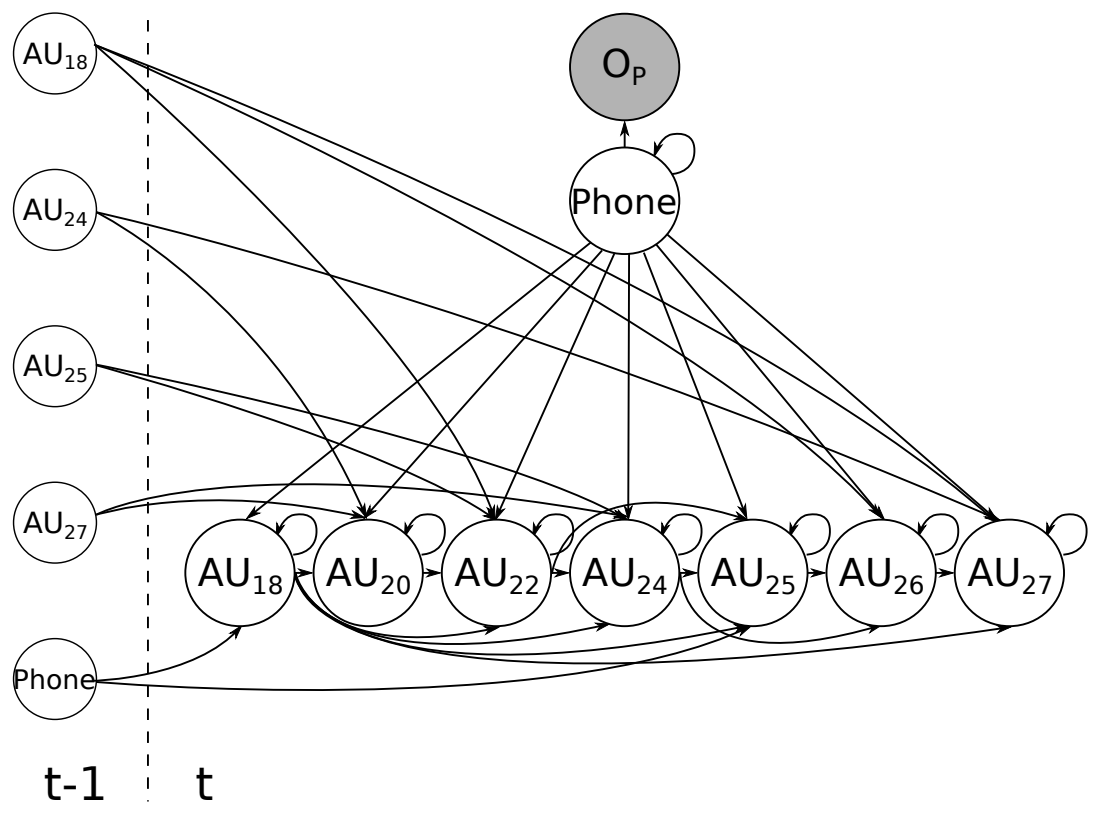

Fig. 9: A DBN model learned from the clean subset for modeling the semantic and dynamic relationships between AUs and phonemes. The directed links in the same time slice represent the semantic relationships among the nodes; the self-loop at each node represents its temporal evolution; and the directed links across two time slices represent the dynamic dependency between the two nodes. The shaded node is the measurement node and employed as evidence for inference; and the unshaded nodes are hidden nodes, whose states can be estimated by inferring over the trained DBN model.

features employed as the input to train an AdaBoost classifier for each AU.

**DBN:** The sixth baseline method, denoted as *DBN*, employs a Dynamic Bayesian Network (DBN) to model the semantic and dynamic relationships between phonemes and AUs. Specifically, the DBN structure as shown in Fig. 9 as well as the DBN parameters are learned using the Bayes Net Toolbox [90] from the clean subset. In order to synchronize phoneme measurements with the image frames, the continuous phoneme segments obtained by speech recognition are discretized according to the sampling rate of the image frames, as illustrated in Fig. 10. Then, AU recognition is performed by DBN inference given the discretized phoneme measurements.

**CTBN-F:** The last baseline method, denoted as *CTBN-F*, which is short for CTBN-Factorized, employs a factorized CTBN to explicitly model the dynamic and physiological relationships between phonemes and each AU as well as the dynamic relationships between AUs. As shown in Fig. 11, each AU is represented by an individual node with 2 states, i.e. "absence" and "presence", in contrast to a combined node in Fig. 5. The directed link between "**Phone**" and each AU node represents the dynamic and physiological relationships between phonemes and the AU. Those between AU nodes capture the dynamic interactions among AUs and are learned from the data using CTBN-lre [85]. The model parameters, i.e., the CIMs, are estimated as described in Section 3.4 from the training data.

Both *DBN* (Fig. 9) and *CTBN-F* (Fig. 11) capture dynamic relationships between AUs and phonemes. However, the dynamic dependencies from AUs to phonemes are not learned and modeled in a *DBN*. This is because the penalty for adding a link from an AU node to the phoneme node is much higher that that from the phoneme node to AU nodes for the 29-state phoneme node. In addition, since loops are allowed in a CTBN model, there is a loop between AU24 and AU25 in *CTBN-F* indicating the strong dynamic relationships between those two AUs.

Note that the *Ada-MFCC*, *SVM-LGBP*, *IB-CNN-LIP*, *DBN*, and *CTBN-F* methods are proposed in this work for recognizing speech-related AUs using only audio information.

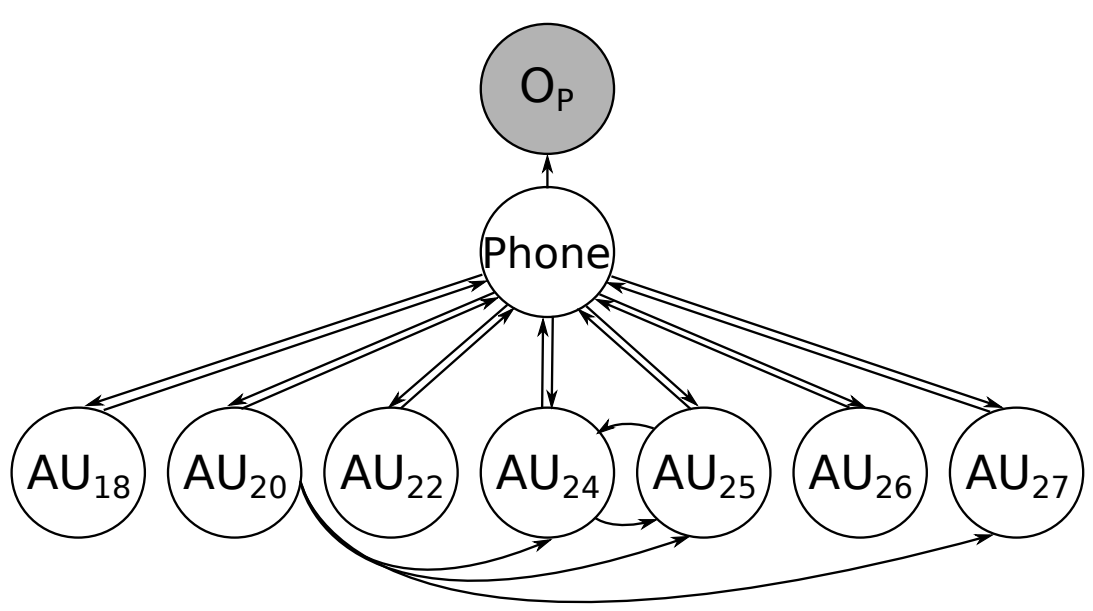

Fig. 11: The structure of a *CTBN-F* model trained on the clean subset for modeling the dynamic physiological relationships between AUs and phonemes.

### 4.3 Experimental Results and Data Analysis on the Clean Subset

We first evaluate the proposed CTBN model on the clean subset. For all methods compared, a leave-one-subject-out training/testing strategy is employed, where the data from 8 subjects is used for training and the remaining data is used for testing. Quantitative experimental results on the clean subset are reported in Fig. 12 in terms of F1 score, true positive rate, and false positive rate. As shown in Fig. 12, the proposed CTBN model outperformed all the baseline methods significantly in terms of the average F1 score (**0.653**).

**Compared with *Ada-LBP*, *Ada-LPQ*, and *SVM-LGBP*,** which employ appearance information from the visual channel, the overall AU recognition performance is improved from **0.416** (*Ada-LBP*), **0.448** (*Ada-LPQ*), and **0.386** (*SVM-LGBP*) to **0.653** by the proposed *CTBN* model in terms of the average F1 score. As shown in Fig. 12, *CTBN* outperforms *Ada-LBP*, *Ada-LPQ*, and *SVM-LGBP* for all target AUs, which demonstrates the effectiveness of using information extracted from the audio channel. The improvement is more impressive for AU27 (mouth stretch), i.e., **0.755** by *CTBN* versus **0.273** by *Ada-LBP*, **0.279** by *Ada-LPQ*, and **0.296** by *SVM-LGBP*, since the visual observation of AU27 is not reliable during speech due to the occlusion caused by lip movements as illustrated in Fig. 1.

**Compared with *IB-CNN-LIP*,** which employs both appearance and geometry information from the visual channel, the overall AU recognition performance is improved from **0.465** (*IB-CNN-LIP*) to **0.653** by the proposed *CTBN* in terms of the average F1 score. Not surprisingly, the *IB-CNN-LIP* outperforms the other visual-based approaches that employ only appearance features. In addition, it also performs better than the proposed *CTBN* on AU25 (lips part) because both the appearance and geometry clues from the visual channel are strong for AU25. In contrast, drastic improvement is achieved for AU26 (jaw drop), from **0.367** by *IB-CNN-LIP* to **0.748** by *CTBN*, because the appearance information for AU26 is invisible due to the occlusion as depicted in Fig. 1 and the geometrical change is subtle during speech.

**Compared with *Ada-MFCC*,** which employs low-level acoustic features in AU recognition, the proposed *CTBN* improves the overall AU recognition performance by **0.217** in terms of the average F1 score. Furthermore, the CTBN outperforms the *Ada-MFCC* for all AUs notably by employing more accurate and reliable higher level audio information, i.e., the phonemes, thanks to the advanced speech recognition techniques, and more importantly, by exploiting the dynamic physiological relationships between AUs and phonemes.

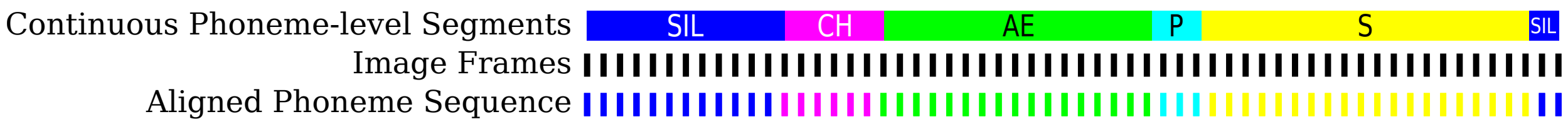

Fig. 10: An illustration of discretizing continuous phoneme segments into frame-by-frame phoneme measurements for the word “chaps”. The first row gives the phoneme-level segments obtained by Kaldi [82]. The second row shows a sequence of image frames, to which the phonemes will be aligned. The last row depicts the aligned sequence of phoneme measurements. Best viewed in color.

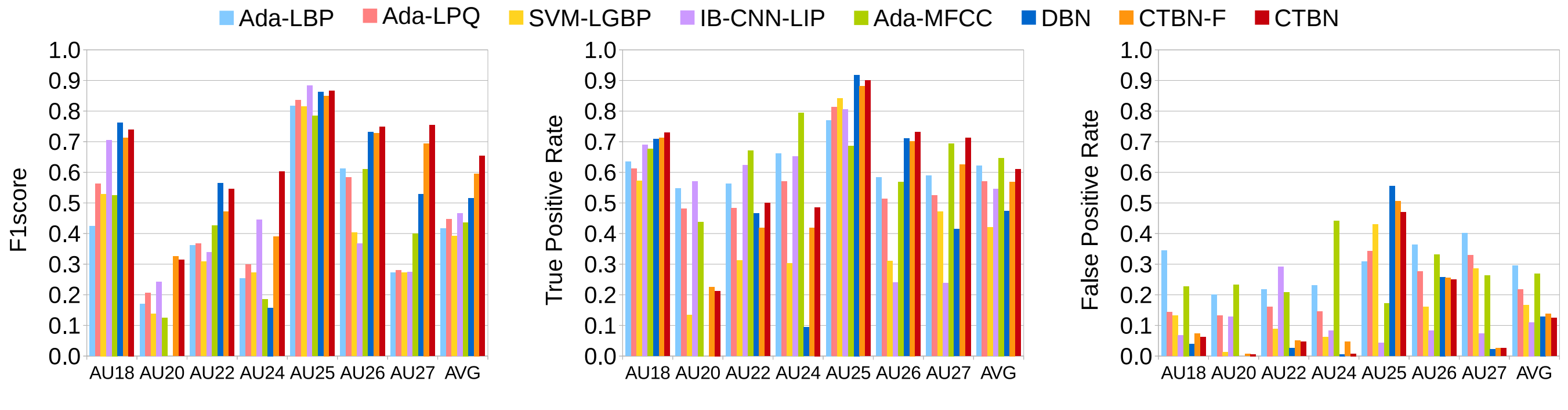

Fig. 12: Performance comparison on the clean subset in terms of (a) F1 score, (b) true positive rate, and (c) false positive rate for the 7 speech-related AUs.

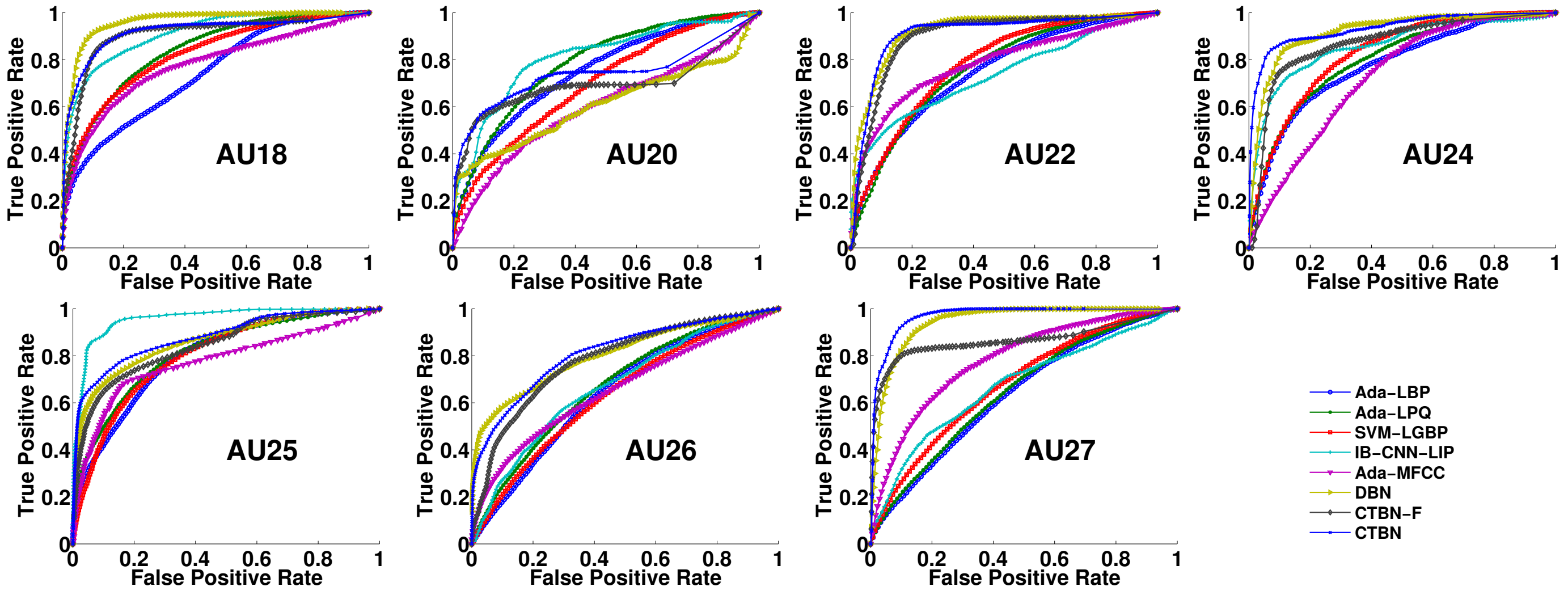

Fig. 13: ROC curves for 7 speech-related AUs on the clean subset.

**Compared with *DBN*,** the proposed *CTBN* improves the overall AU recognition performance by **0.138**, in terms of the average F1 score. Particularly, the F1 scores of *CTBN* are better than or at least comparable to those of *DBN* for all AUs, as shown in Fig. 12. The primary reason for the performance improvement is that the dynamic dependencies modeled in *DBN* are stationary; whereas the relationships between AUs and phonemes actually undergo a temporal evolution as modeled in the *CTBN*. For example, the F1 score of AU24 (lip presser) is dramatically improved from **0.158** by *DBN* to **0.603** by *CTBN*, because AU24 is activated before the sound is produced and released once the sound is heard, which can be better modeled in *CTBN*. Note that *DBN* fails to recognize AU20 (lip stretcher). Although AU20 is required to produce AE in ch***a***ps according to Phonetics [78], some subjects did not activate AU20 as observed in our audiovisual dataset and thus, the semantic relationship between **Phone** and **AU20** is rather weak. However, no dynamic link is learned between **Phone** and **AU20** in *DBN*. In contrast, dynamic relationships between AUs and phonemes modeled by *CTBN* are more crucial for inferring AU20. As a result, the F1 score of AU20 is improved from **0** by *DBN* to **0.314** by *CTBN*. We found that *DBN* performs slightly better on AU18 (lip pucker) and AU22 (lip funneler) than *CTBN*. This is because AU18 and AU22 have the strongest static relationships with phonemes: when pronouncing UW in t***wo*** and CH in ***ch***eese, they are activated for most of subjects.

**Compared with *CTBN-F*,** the proposed *CTBN* further improves the overall AU recognition performance by **0.057**, in terms of the average F1 score. By employing one single node to model the joint distribution over the 7 target AUs, comprehensive relationships between AUs and phonemes, i.e., AUs occur in combinations to produce sounds, can be well characterized as discussed in Section 3.2.

In addition, Fig. 14 gives an example of the system outputs of the *CTBN* and *CTBN-F*, i.e., the probabilities of AUs given the phoneme measurements (the shaded phoneme sequence), by the CTBN inference over continuous time. For both *CTBN* and *CTBN-F*, the probabilities of AUs change corresponding to the transitions of phonemes, when sounding a word “beige”. For example, the probability of AU24 increases and reaches its apex before the phoneme *B* is produced. AU26 can be recognized even though the gap between upper and lower teeth is invisible in visual channel because the presence of AU24. When the sound *B* is emitted, the probability of AU24 drops rapidly, while the probability of AU25 increases. The vertical green line denotes the time point when AU24 is released, while the vertical garnet line denotes the time point when AU25 is activated. Ideally, they should overlap with each other due to the transition from the "Stop" phase to the "Aspiration" phase of sounding *B*, as the result of the *CTBN* (the top plot). Whereas, a noticeable gap between the two lines can be observed in the result of the *CTBN-F* (the bottom plot) because that no two AUs are allowed to change states at the same time in the factorized CTBN.

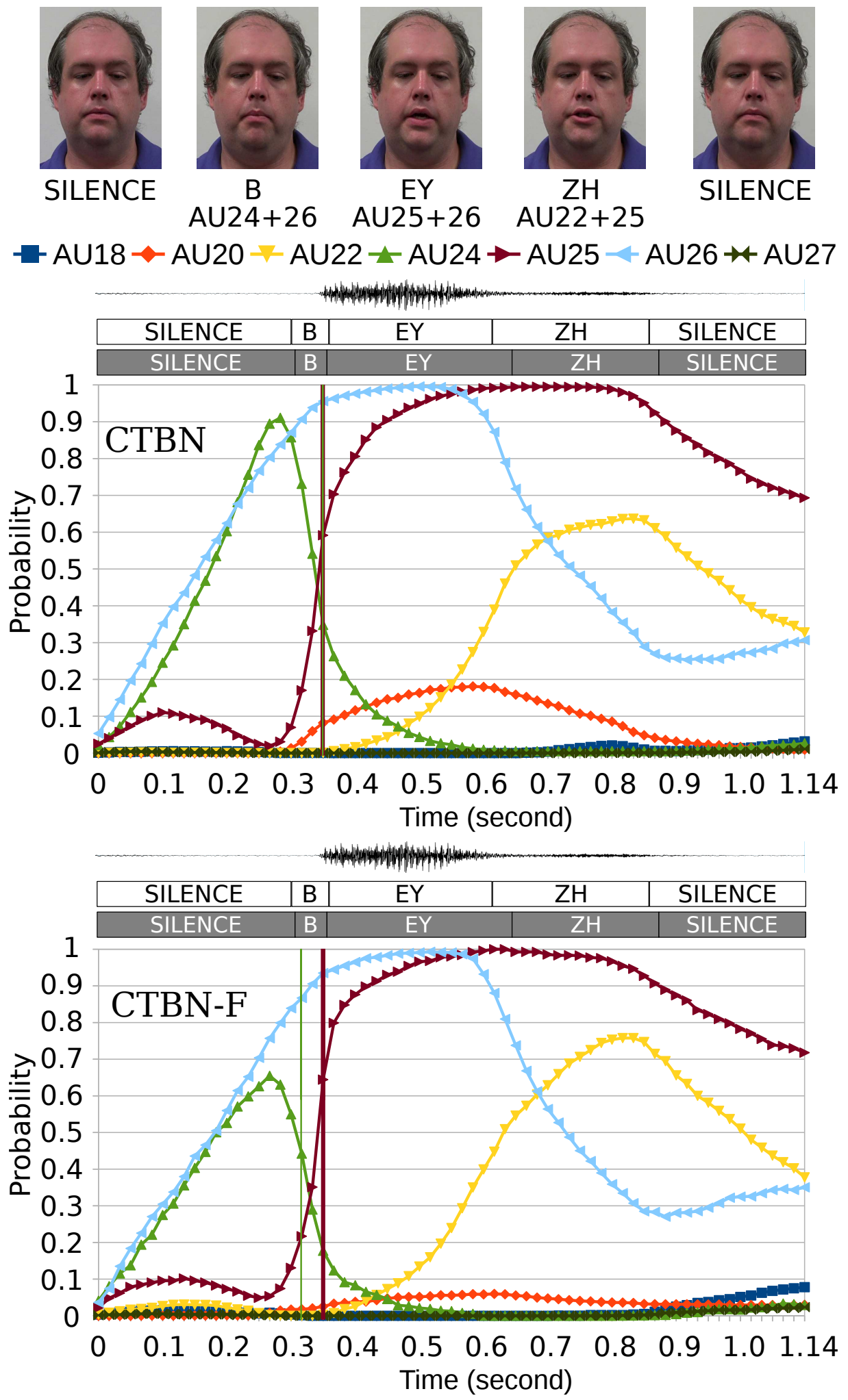

Fig. 14: An example of the system outputs by CTBN inference using *CTBN* and *CTBN-F*, respectively. The top row shows key frames from an image sequence where a word "beige" is produced, where AU22, AU24, AU25, and AU26 are involved. The bottom two figures depict the probabilities of AUs changing over time by *CTBN* and *CTBN-F*, respectively. The shaded phoneme sequence is used as evidence of the CTBN models and the unshaded one is the ground truth phoneme labels. The vertical green line denotes the time point when AU24 is released, while the vertical garnet line denotes the time point when AU25 is activated. The two lines are overlapped with each other in the *CTBN* output. Best viewed in color.

Moreover, we analyzed the relationships learned by *CTBN*. Table 2 depicts a part of the CIM associated with the "**AU**" node given the state of "**Phone**" as *B*, where the first row and column give the states of "**AU**" with the corresponding AU/AU combinations. For example, if "**AU**" is at its $10^{th}$ state, i.e., AU24 and AU26 are activated, the corresponding conditional intensity value in the CIM is $-16.72$. As described in Section 3.3, the "**AU**" node is expected to transit in $\frac{1}{16.72}s$. Upon transition, it has a higher chance to transit to its $6^{th}$ state (AU25+AU26) with a probability of $\frac{9.12}{16.72}$. However, if the lip movement is not fast, i.e., AU25 is not activated when AU24 is released, it may transit to its $2^{nd}$ state (AU26) with a probability of $\frac{6.59}{16.72}$. Then, the "**AU**" node will leave this state quickly in $\frac{1}{40.47}s$ and transit to its $6^{th}$ state with a high probability of $\frac{32.37}{40.47}$.

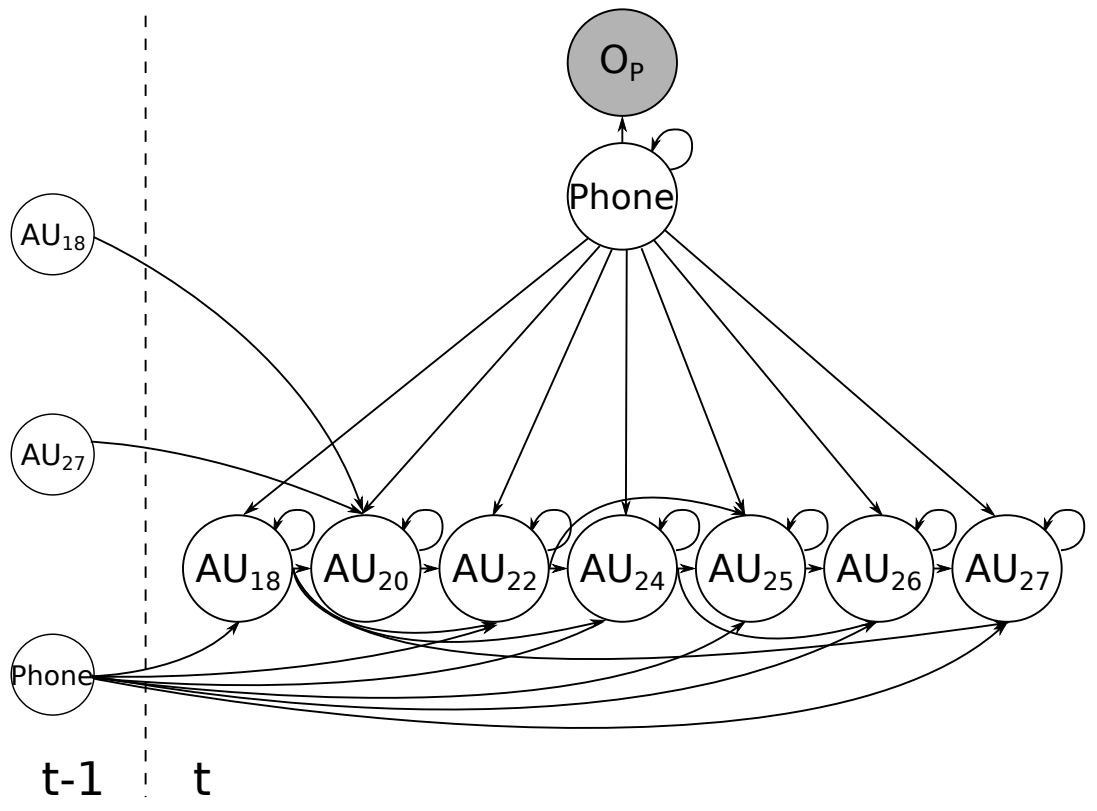

Fig. 15: A DBN model learned from the challenging subset for modeling the semantic and dynamic relationships between AUs and phonemes.

TABLE 2: A part of the CIM associated with the "**AU**" node given the state of "**Phone**" as *B*, where the first row and column give the states of "**AU**" node with the corresponding AU/AU combinations in the parenthesis.

| | 0 | ··· | 2(AU26) | ··· | 6(AU25+AU26) | ··· | 10(AU24+AU26) | ··· |
|---|---|---|---|---|---|---|---|---|
| 0 | −10.07 | ··· | 0 | ··· | 0 | ··· | 0 | ··· |
| ⋮ | ⋮ | ⋮ | ⋮ | ⋮ | ⋮ | ⋮ | ⋮ | ⋮ |
| 2 | 0 | ··· | −40.47 | ··· | 32.37 | ··· | 8.09 | ··· |
| ⋮ | ⋮ | ⋮ | ⋮ | ⋮ | ⋮ | ⋮ | ⋮ | ⋮ |
| 6 | 1.23 | ··· | 0 | ··· | −1.23 | ··· | 0 | ··· |
| ⋮ | ⋮ | ⋮ | ⋮ | ⋮ | ⋮ | ⋮ | ⋮ | ⋮ |
| 10 | 0 | ··· | 6.59 | ··· | 9.12 | ··· | −19.86 | ··· |
| ⋮ | ⋮ | ⋮ | ⋮ | ⋮ | ⋮ | ⋮ | ⋮ | ⋮ |

We further performed a Receiver Operating Characteristic (ROC) analysis for each AU. An ROC curve is obtained by plotting the true positive rates against the false positive rates while varying the decision threshold of the predicted scores. As shown in Fig. 13, the performance of the proposed *CTBN* is better than or at least comparable to that of the baseline methods for all the target AUs. The improvement is more significant on the AUs that have strong dynamic relationships with phonemes, i.e. AU24, or whose visual observation is not reliable during speech, i.e. AU27.

### 4.4 Experimental Results on the Challenging Subset of the Audiovisual Database

Experiments were conducted on the challenging subset to demonstrate the effectiveness of the proposed audio-based facial AU recognition under real world conditions, where facial activities are accompanied by free head movements, illumination changes, and often with occlusions of the face regions caused by facial hairs, caps, or glasses.

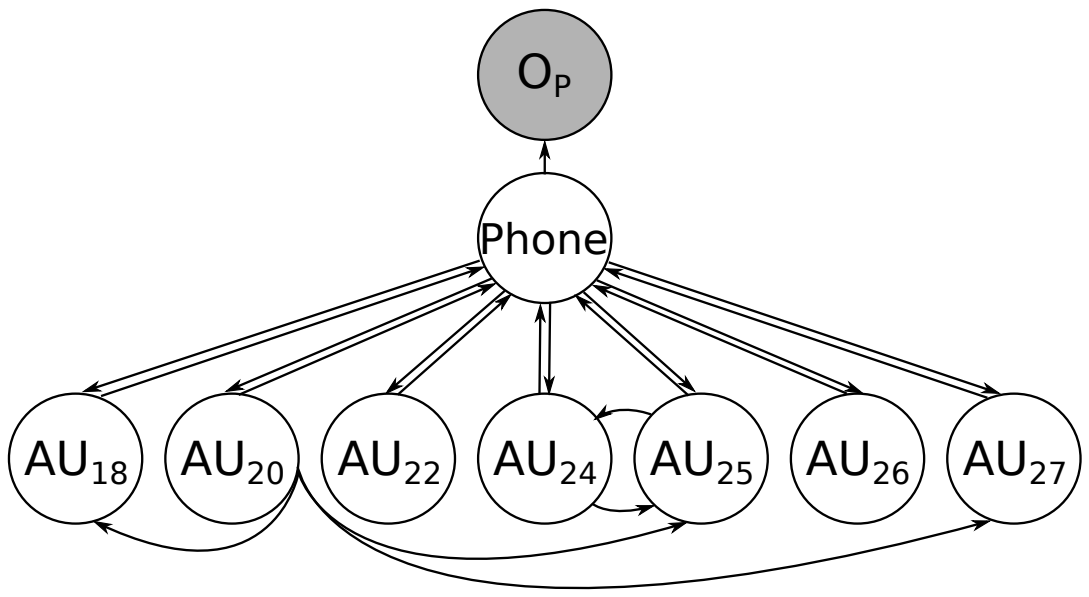

Fig. 16: A *CTBN-F* model trained on the challenging subset for modeling the dynamic physiological relationships between AUs and phonemes.

The proposed *CTBN* model and the baseline methods, i.e. *Ada-LBP*, *Ada-LPQ*, *SVM-LGBP*, *IB-CNN-LIP*, *Ada-MFCC*, *DBN*, and *CTBN-F*, were trained and tested on the challenging subset using a leave-one-subject-out strategy. Since there are only 6 subjects in the challenging subset, we employed the data in the clean subset except those of the two subjects, who also appear in the challenging subset, as additional training data to ensure a subject-independent context. Specifically, the data of 5 subjects from the challenging subset along with the data of 7 subjects from the clean subset is used as the training data, and the remaining one subject from the challenging subset is employed as the testing data.

The structures of the *DBN* and *CTBN-F* trained on the challenging data are shown in Fig. 15 and Fig. 16, respectively. Comparing Fig. 9 and Fig. 15, we can see that the dynamic relationships from phonemes to AUs become more important on the challenging subset in the *DBN* model, i.e. more temporal links are learned from the phoneme node of the $t-1^{th}$ slice to AU nodes of the $t^{th}$ slice in Fig. 15. This is because the labeling uncertainty of AUs is alleviated in the challenging subset, especially for non-frontal faces, since lip movement is often asymmetrical during speech [7].

### 4.4.1 *Experimental Results and Discussion*

Quantitative experimental results are reported in Fig. 17 in terms of F1 score, true positive rate, and false positive rate. As shown in Fig. 17, the proposed *CTBN* achieved the best recognition performance among all the methods compared, in terms of the average F1 score (**0.682**).

Note that the performance of the visual-based methods degrades significantly on the challenging subset even with more training data. As shown in Table 3, the average F1 score of *Ada-LBP* decreases from **0.416** (clean) to **0.372** (challenging); that of *Ada-LPQ* decreases from **0.448** (clean) to **0.362** (challenging); that of *SVM-LGBP* decreases from **0.386** (clean) to **0.339** (challenging); and that of *IB-CNN-LIP* drops from **0.465** to **0.382** due to large face pose variations and occlusions on the face regions. In contrast, the information extracted from the audio channel is robust to head movements and occlusions for facial AU recognition. As a result, the performance of the audio-based methods on the challenging subset is comparable or even slightly better than that on the clean subset because of employing additional training data, as reported in Table 3.

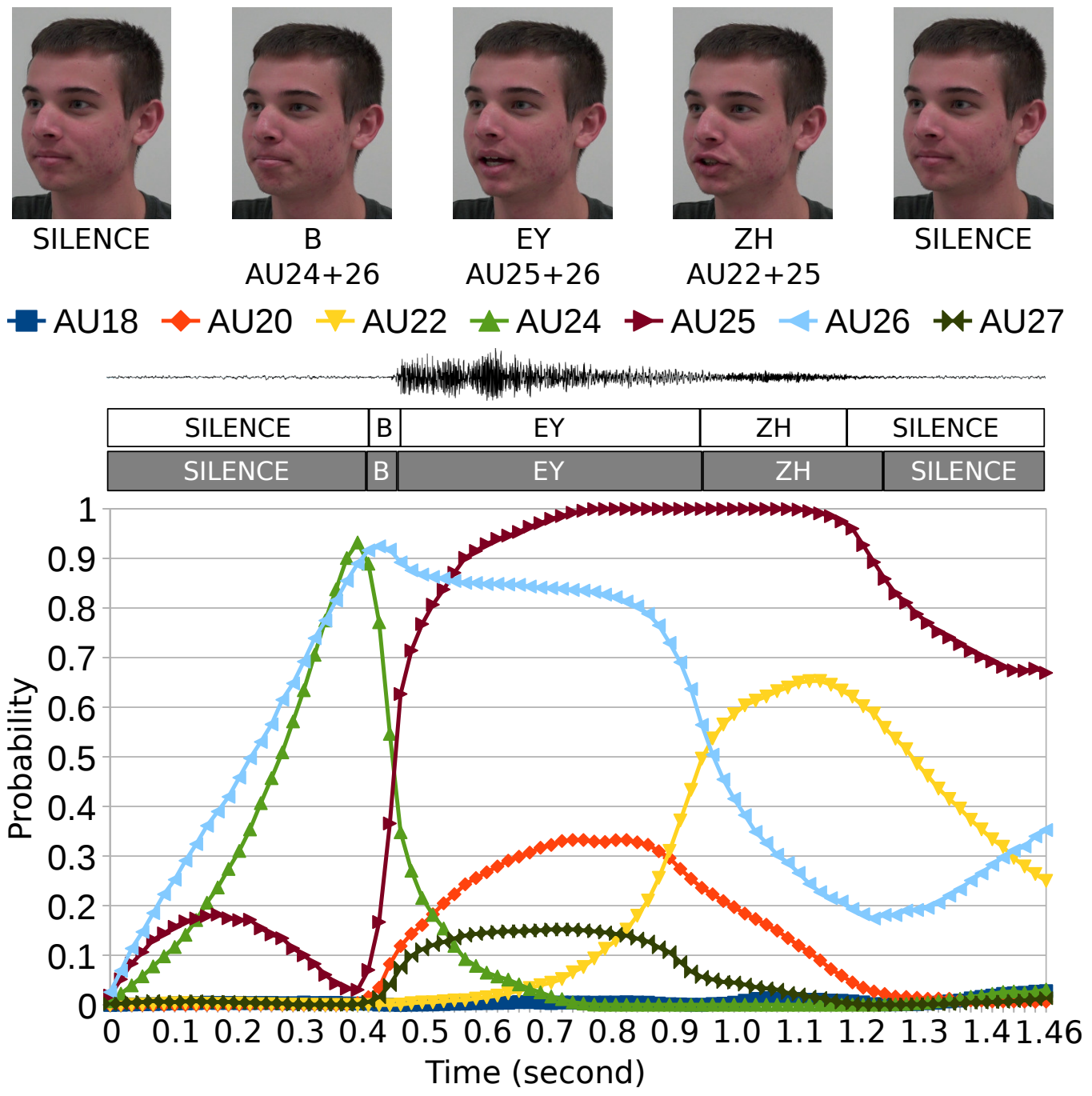

Fig. 18: An example of the system outputs by *CTBN* inference on the challenging subset. The top row shows key frames from an image sequence where a word "beige" is produced and AU22, AU24, AU25, and AU26 are involved. The bottom figure depicts the probabilities of AUs changing over time. The shaded phoneme sequence is used as evidence of the *CTBN* and the unshaded one is the ground truth phoneme labels.. Best viewed in color.

TABLE 3: Performance comparison on the two subsets in terms of the average F1 score.

| **Subsets** | *Ada-LBP* | *Ada-LPQ* | *SVM-LGBP* | *IB-CNN-LIP* | *Ada-MFCC* | *DBN* | *CTBN-F* | *CTBN* |
|---|---|---|---|---|---|---|---|---|
| **Clean** | 0.416 | 0.448 | 0.386 | 0.465 | 0.436 | 0.515 | 0.596 | **0.653** |
| **Challenging** | 0.372 | 0.362 | 0.339 | 0.382 | 0.445 | 0.534 | 0.589 | **0.682** |

An ROC analysis was performed for each target AU on the challenging subset. As shown in Fig. 19, the performance of the proposed *CTBN* is better or at least comparable to that of the baseline methods for all the target AUs. In addition, an example of the system outputs of the *CTBN* over continuous time is illustrate in Fig. 18. As depicted by Fig. 18, the probabilities of AUs undergo a similar temporal evolution corresponding to the transitions of phonemes as the case in the clean subset.

## 4.5 Analysis on Phoneme Measurement

The proposed audiovisual fusion framework benefits from the remarkable achievements in speech recognition. In our experiments, the speech recognition performance of the Kaldi Toolkit [82] is 1.3% (7/540) on the clean subset and 1.4% (5/360) on the challenging subset in terms of the word-level error rate (WER, [insert+delete+substitute]/[number of words]). To evaluate the effect of phoneme measurement on fusion, we have conducted an experiment using the ground-truth phoneme segments as the evidence for the CTBN model, denoted as *CTBN-perfect*. As shown in Fig. 20, *CTBN* using phoneme measurements from speech recognition yields comparable performance with *CTBN-perfect* using ground-truth phoneme segments.

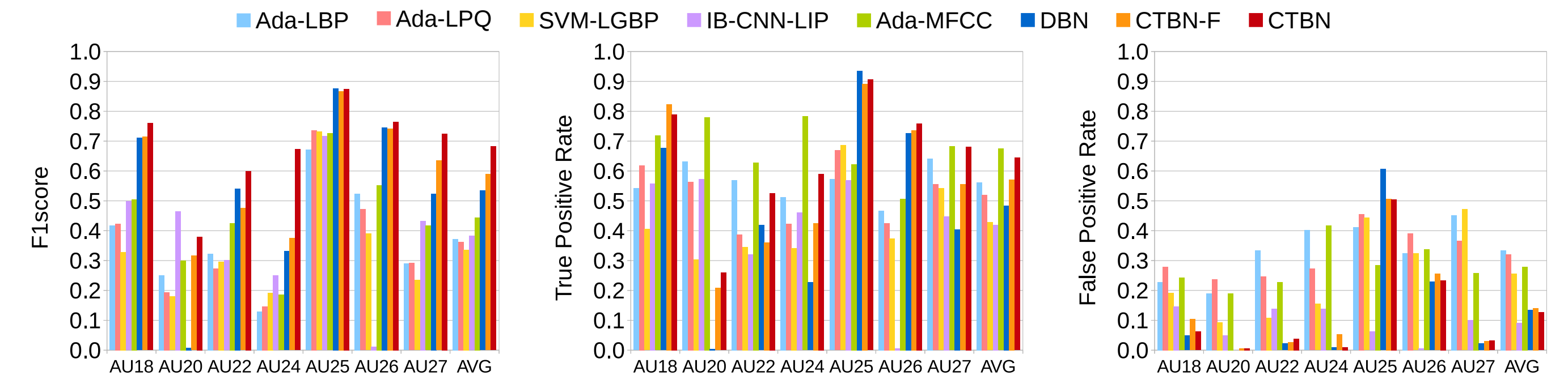

(a) (b) (c)

Fig. 17: Performance comparison on the challenging subset in terms of (a) F1 score, (b) true positive rate, and (c) false positive rate for the 7 speech-related AUs.

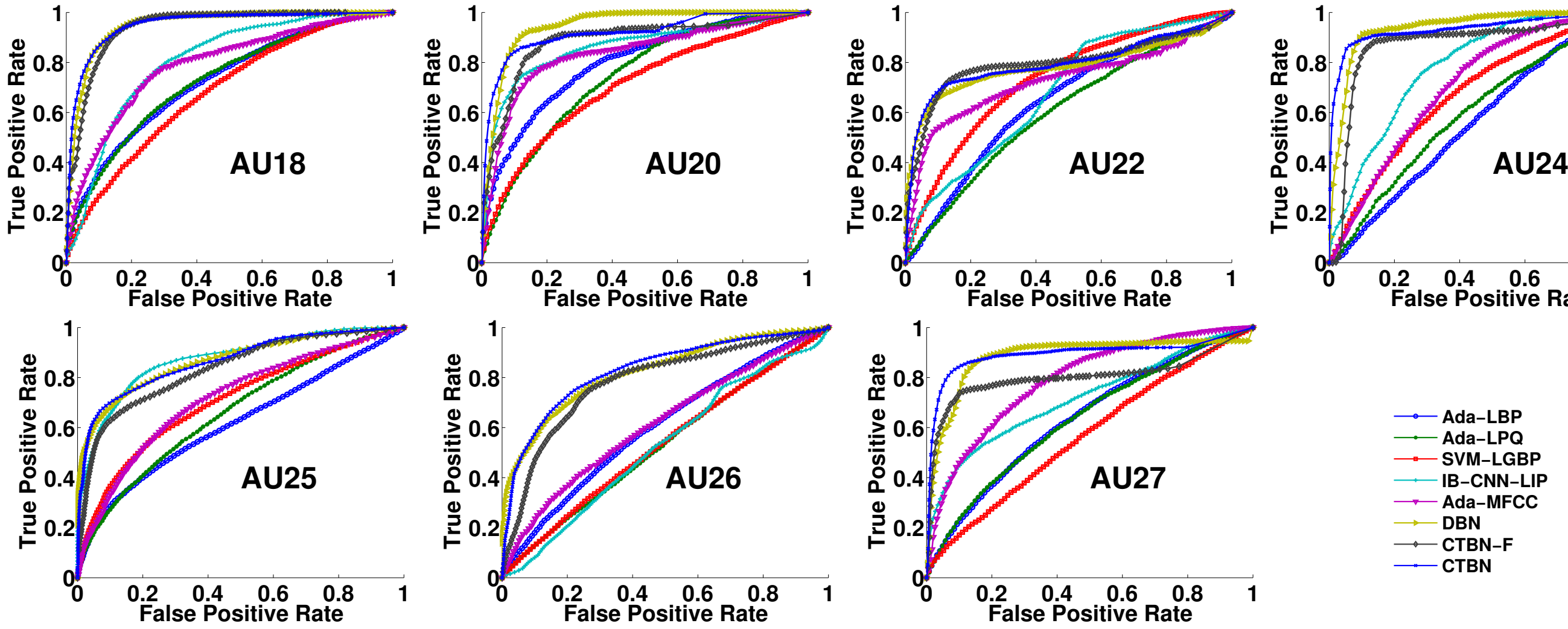

Fig. 19: ROC curves for 7 speech-related AUs on the challenging subset.

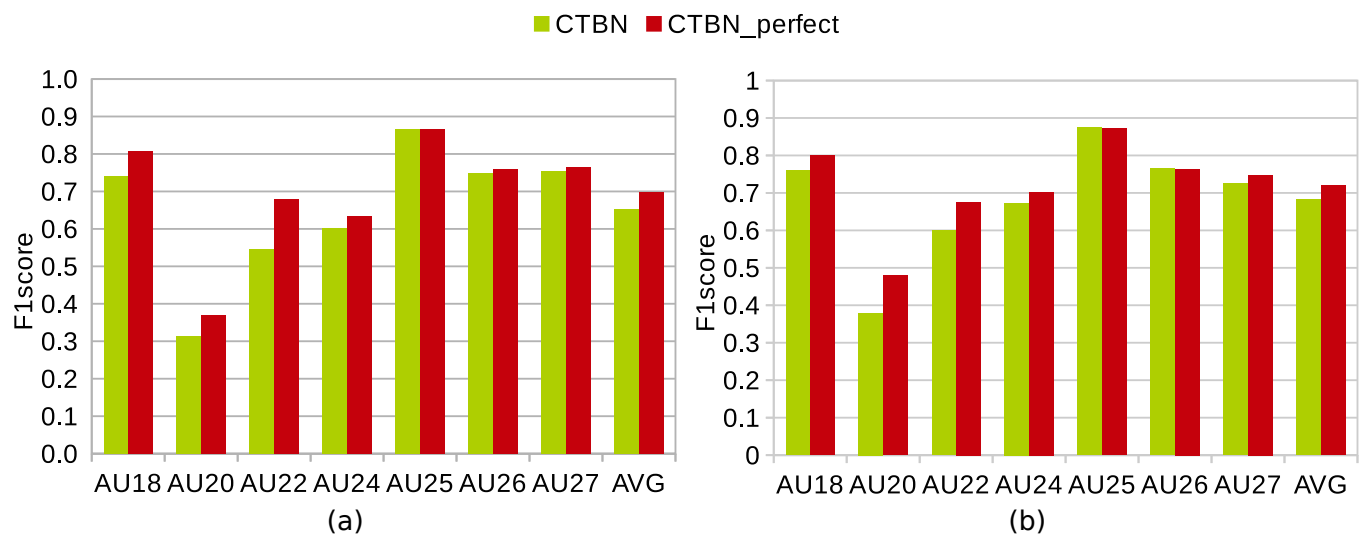

Fig. 20: Performance comparison between *CTBN* and *CTBN-perfect* in terms of F1 score on (a) the clean subset and (b) the challenging subset.

## 5 Conclusion and Future Work

It is challenging to recognize speech-related AUs due to the subtle facial appearance and geometrical changes as well as occlusions introduced by frequent lip movements. In this work, we proposed a novel audio-based AU recognition framework by exploiting information from the audio channel, i.e., phonemes in particular, because facial activities are highly correlated with voice. Specifically, a CTBN model is employed to model the dynamic and physiological relationships between phonemes and AUs, as well as the temporal evolution of these relationships. Given the phoneme measurements, AU recognition is then performed by probabilistic inference through the CTBN model.

Experimental results on a new audiovisual AU-coded dataset have demonstrated that the CTBN model achieved significant improvement over the state-of-the-art visual-based AU recognition method. The improvement is more impressive for those AUs, whose visual observations are impaired during speech. More importantly, the experimental results on the challenging subset have demonstrated the effectiveness of utilizing audio information for recognizing speech-related AUs under real world conditions, where the visual observations are not reliable. Furthermore, the proposed CTBN model also outperformed the other baseline methods employing audio signals, thanks to explicitly modeling the dynamic interactions between phonemes and AUs in the context of human communication.

In the future, we plan to extend the audiovisual database to include continuous and emotional speech, while extensive labeling workload is expected for AU annotation. The framework learned from the enriched database is expected to capture more comprehensive relationships in natural human communication. Under the contexts of emotion, more AUs especially the upper-face AUs can be modeled. In addition, it is expected to be more robust to imperfect phoneme measurements by modeling the relationships in the words.

## Acknowledgments

This work is supported by National Science Foundation under CAREER Award IIS-1149787.

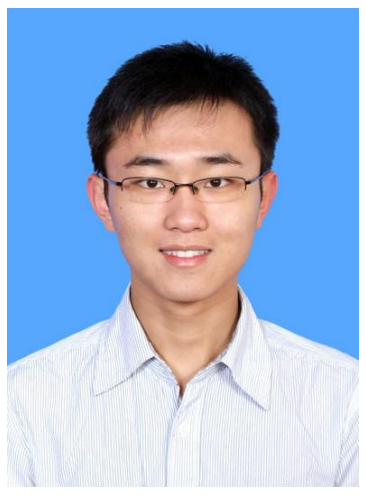

**Zibo Meng** received a Master degree from Zhejiang University, China, in 2013. He is currently pursuing his Ph.D. degree at University of South Carolina, Columbia, South Carolina. His areas of research include computer vision, pattern recognition, and information fusion. He is a student member of the IEEE.

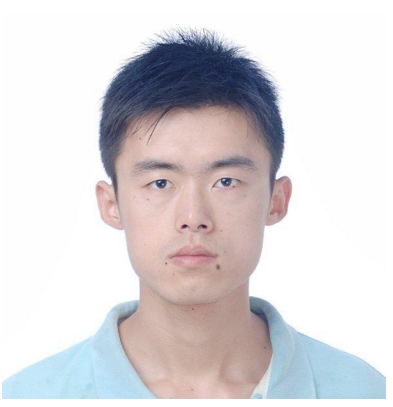

**Shizhong Han** is pursuing his Ph.D. degree at University of South Carolina, Columbia, South Carolina. He received a Master degree from Wuhan University, China, in 2010. His research interests include computer vision, pattern recognition, machine learning, and data mining. He is a student member of the IEEE.

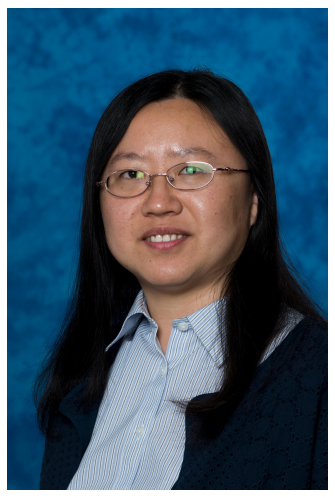

**Yan Tong** received the Ph.D. degree in electrical engineering from Rensselaer Polytechnic Institute, Troy, New York, in 2007. She is currently an associate professor in the Department of Computer Science and Engineering, University of South Carolina, Columbia, SC, USA. From 2008 to 2010, she was a research scientist in the Visualization and Computer Vision Lab of GE Global Research, Niskayuna, NY. Her research interests include computer vision, machine learning, and human computer interaction. She has served as a conference organizer and a program committee member for a number of premier international conferences. She is a member of the IEEE.